\documentclass{article}
\usepackage[utf8]{inputenc}
\usepackage[T1]{fontenc}
\usepackage{amsmath,amssymb}
\usepackage{graphicx}
\usepackage{booktabs}
\usepackage[numbers]{natbib}
\usepackage{hyperref}
\usepackage{geometry}
\usepackage{xcolor}
\usepackage{multirow}
\usepackage{array}
\usepackage{caption}
\usepackage{float}

\begin{document}

\title{Agentic Automation of BT-RADS Scoring: End-to-End Multi-Agent System for Standardized Brain Tumor Follow-up Assessment}

\author{
Mohamed Sobhi Jabal, MD$^{1}$ \and
Jikai Zhang, MS$^{2,3}$ \and
Dominic LaBella, MD$^{4}$ \and
Jessica L. Houk, MD$^{1}$ \and
Dylan Zhang, MD$^{1,7}$ \and
Jeffrey D. Rudie, MD, PhD$^{5,8}$ \and
Kirti Magudia, MD, PhD$^{1}$ \and
Maciej A. Mazurowski, PhD$^{1,2,6}$ \and
Evan Calabrese, MD, PhD$^{1,3}$
}

\date{}

\maketitle

\noindent
$^{1}$Department of Radiology, Duke University Medical Center, Durham, NC\\
$^{2}$Department of Electrical and Computer Engineering, Duke University, Durham, NC\\
$^{3}$Duke Center for Artificial Intelligence in Radiology, Duke University Medical Center, Durham, NC\\
$^{4}$Department of Radiation Oncology, Duke University Medical Center, Durham, NC\\
$^{5}$Department of Radiology, University of California San Diego, San Diego, CA\\
$^{6}$Department of Biostatistics and Bioinformatics, Duke University School of Medicine, Durham, NC\\
$^{7}$Department of Radiology, Santa Clara Valley Medical Center, San Jose, CA\\
$^{8}$Department of Radiology, Scripps Clinic Medical Group, San Diego, CA

\vspace{1em}

%% ============================================================
%% ABSTRACT
%% ============================================================
\begin{abstract}

\noindent\textbf{Purpose:}
The Brain Tumor Reporting and Data System (BT-RADS) standardizes post-treatment MRI response assessment in patients with diffuse gliomas but requires complex integration of imaging trends, medication effects, and radiation timing. This study evaluates an end-to-end multi-agent large language model (LLM) and convolutional neural network (CNN) system for automated BT-RADS classification.

\noindent\textbf{Materials and Methods:}
A multi-agent LLM system combined with automated CNN-based tumor segmentation was retrospectively evaluated on 509 consecutive post-treatment glioma MRI examinations from a single high-volume center. An extractor agent identified clinical variables (steroid status, bevacizumab status, radiation date) from unstructured clinical notes, while a scorer agent applied BT-RADS decision logic integrating extracted variables with volumetric measurements. Expert reference standard BT-RADS classifications were established by an independent board-certified neuroradiologist. Initial clinical assessments from routine radiology reports served as comparators.

\noindent\textbf{Results:}
Of 509 examinations, 492 met inclusion criteria. The system achieved 374/492 (76.0\%; 95\% CI, 72.1\%--79.6\%) accuracy versus 283/492 (57.5\%; 95\% CI, 53.1\%--61.8\%) for initial clinical assessments (+18.5 percentage points; $P <.001$). Context-dependent categories showed high sensitivity (BT-1b 100\%, BT-1a 92.7\%, BT-3a 87.5\%), while threshold-dependent categories showed moderate sensitivity (BT-3c 74.8\%, BT-2 69.2\%, BT-4 69.3\%, BT-3b 57.1\%). For BT-4, positive predictive value was 92.9\%.

\noindent\textbf{Conclusion:}
The multi-agent LLM system achieved higher BT-RADS classification agreement with reference standard scoring compared to initial clinical scoring. We observed high accuracy for clinical context-dependent scores, moderate accuracy for volumetric threshold-dependent scores, and high positive predictive value for BT-4. These findings support the potential of multi-agent architectures for automating standardized clinical assessment frameworks in neuro-oncology.

\vspace{0.5em}
\end{abstract}

%% ============================================================
%% INTRODUCTION
%% ============================================================
\section{Introduction}

Glioblastoma is the most common malignant primary brain tumor, with median survival of approximately 15 months despite maximal multimodal therapy~\cite{ostrom2021,tan2020}. Tumor recurrence is nearly universal, with 80--90\% occurring at or near the original tumor site~\cite{tan2020}. Differentiating true progression from treatment-related changes such as pseudoprogression and radiation necrosis remains a critical challenge, as misinterpretation may lead to premature cessation of effective therapy or unnecessary intervention for treatment-related changes~\cite{brandsma2008,abbasi2018}.

Structured reporting frameworks have transformed radiology interpretation. The Breast Imaging Reporting and Data System (BI-RADS), introduced in 1993, demonstrated that standardized lexicons and categorical assessments improve communication and outcome tracking~\cite{dorsi2013}. The Brain Tumor Reporting and Data System (BT-RADS) extends this approach to post-treatment glioma surveillance, providing a management-based categorical assessment from BT-0 (not scorable) through BT-4 (high suspicion for tumor progression) that integrates volumetric changes, medication effects, and radiation timing~\cite{weinberg2018,btrads2026}. Institutional implementation has shown improved reporting consistency and referring physician comprehension~\cite{gore2019}, and an interrater agreement study demonstrated good reliability for BT-RADS scoring~\cite{essien2024}.

Artificial intelligence applications in radiology have expanded substantially, with over 70\% of FDA-cleared AI devices targeting imaging~\cite{zhu2022} and NIH funding for AI in radiology increasing 13.7-fold between 2015 and 2024, with AI projected to reach the trillion-dollar economic threshold within the next two decades~\cite{jabal2026}. For brain tumors, AI applications span user-centered decision support systems~\cite{prince2024} to emerging agentic architectures~\cite{dietrich2025}, while deep learning segmentation has advanced from early benchmarks~\cite{menze2015} through multi-parametric approaches~\cite{ghadimi2025} to current generation methods with clinical applicability~\cite{diana2025}.

More recently, large language models have demonstrated strong performance for clinical information extraction from unstructured text~\cite{gu2025,reichenpfader2024}. A practical primer by Bhayana provides comprehensive guidance on LLM applications in radiology~\cite{bhayana2024}. Recent studies have applied LLMs specifically to extraction tasks in radiology reports~\cite{li2025}, though systematic reviews note performance variability and implementation challenges that require careful validation~\cite{keshavarz2024,artsi2025,lee2025scoping}.

Multi-agent LLM architectures, comprising specialized AI agents coordinating across tasks, have emerged as a paradigm for complex healthcare applications~\cite{borkowski2025,ke2024}. Foundational work by Liu and colleagues establishes frameworks for planning, action, and observation cycles in healthcare AI agents~\cite{liu2025}. This builds upon broader developments in medical AI, where LLMs have shown potential across clinical knowledge tasks~\cite{thirunavukarasu2023,singhal2023} and multi-agent conversations can enhance diagnostic reasoning~\cite{chen2025}. In oncology specifically, autonomous AI agents have demonstrated promise in clinical decision-making workflows~\cite{ferber2025}.

Despite these parallel advances in LLMs and automated volumetrics, integration of these technologies for automated treatment response classification remains unexplored. Lee and colleagues demonstrated feasibility of NLP-based BT-RADS inference from unstructured MRI reports~\cite{lee2020}, but relied solely on report text without imaging data. Studies on open-weight LLMs for structured extraction from radiology reports have established extraction methodology applicable to this domain~\cite{leguellec2024,jabal2025}, while automated segmentation has shown promise for volumetric assessment~\cite{zhang2025}. Our study integrates these approaches into an end-to-end system combining automated tumor volumetrics, LLM-based clinical variable extraction, and algorithmic BT-RADS scoring.

%% ============================================================
%% MATERIALS AND METHODS
%% ============================================================
\section{Materials and Methods}

\subsection{Study Design and Cohort}

The institutional review board approved this retrospective study, and informed consent was waived given the nature of the study.

We conducted a retrospective evaluation of a multi-agent LLM and CNN system for automated BT-RADS classification at a single academic institution. The cohort comprised 509 consecutive post-treatment glioma MRI examinations at a single high-volume brain tumor center. Specific data analyzed for each examination included: free-text clinical notes providing medication and treatment history, and MRI sequences used for automated volumetric measurement. Imaging data were collected as part of routine clinical care using the institution's standardized brain tumor MRI protocol~\cite{ellingson2015}.

\subsection{Automated Volumetric Tumor Measurements}

Volumetric tumor measurements were obtained using a previously described and validated deep learning segmentation algorithm~\cite{zhang2025}. A 3D convolutional neural network implemented in nnU-Net was trained on manually annotated post-treatment glioma MRI examinations to automatically segment four tumor compartments: enhancing tumor, non-enhancing tumor, peritumoral edema (FLAIR hyperintensity), and surgical cavity. For each examination, total FLAIR volume (sum of all non-enhancing abnormal regions) and total enhancement volume were calculated from the segmentation outputs. Percentage volume change between baseline and follow-up examinations was computed for both FLAIR and enhancement compartments and used as input features for BT-RADS classification.

\subsection{Multi-Agent System Architecture}

\begin{figure}[H]
\centering
\includegraphics[width=\textwidth]{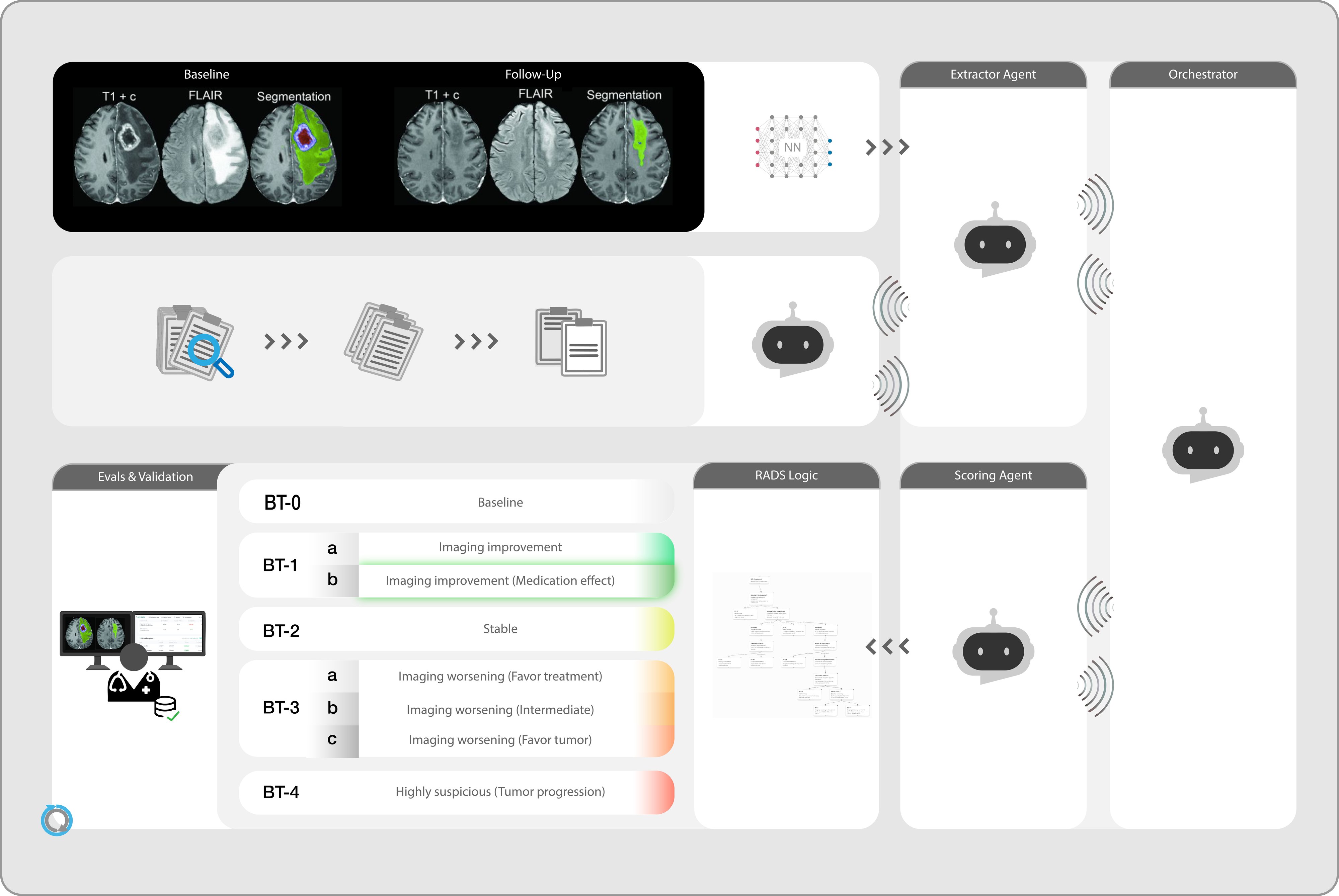}
\caption{Overview of the multi-agent system for automated BT-RADS classification. The pipeline integrates CNN-based tumor segmentation (nnU-Net) for volumetric quantification of FLAIR and enhancement volumes with a 20-billion parameter open-weight LLM (extractor agent) that identifies clinical variables (steroid status, bevacizumab use, radiation date) from unstructured clinical notes with evidence span linking, and a scorer agent that applies BT-RADS decision logic integrating extracted variables with quantitative volumetric measurements. An Orchestrator coordinates data flow between agents. Schema-constrained generation with Pydantic validation ensures outputs conform to predefined formats.}
\label{fig:architecture}
\end{figure}

\begin{figure}[H]
\centering
\includegraphics[width=\textwidth]{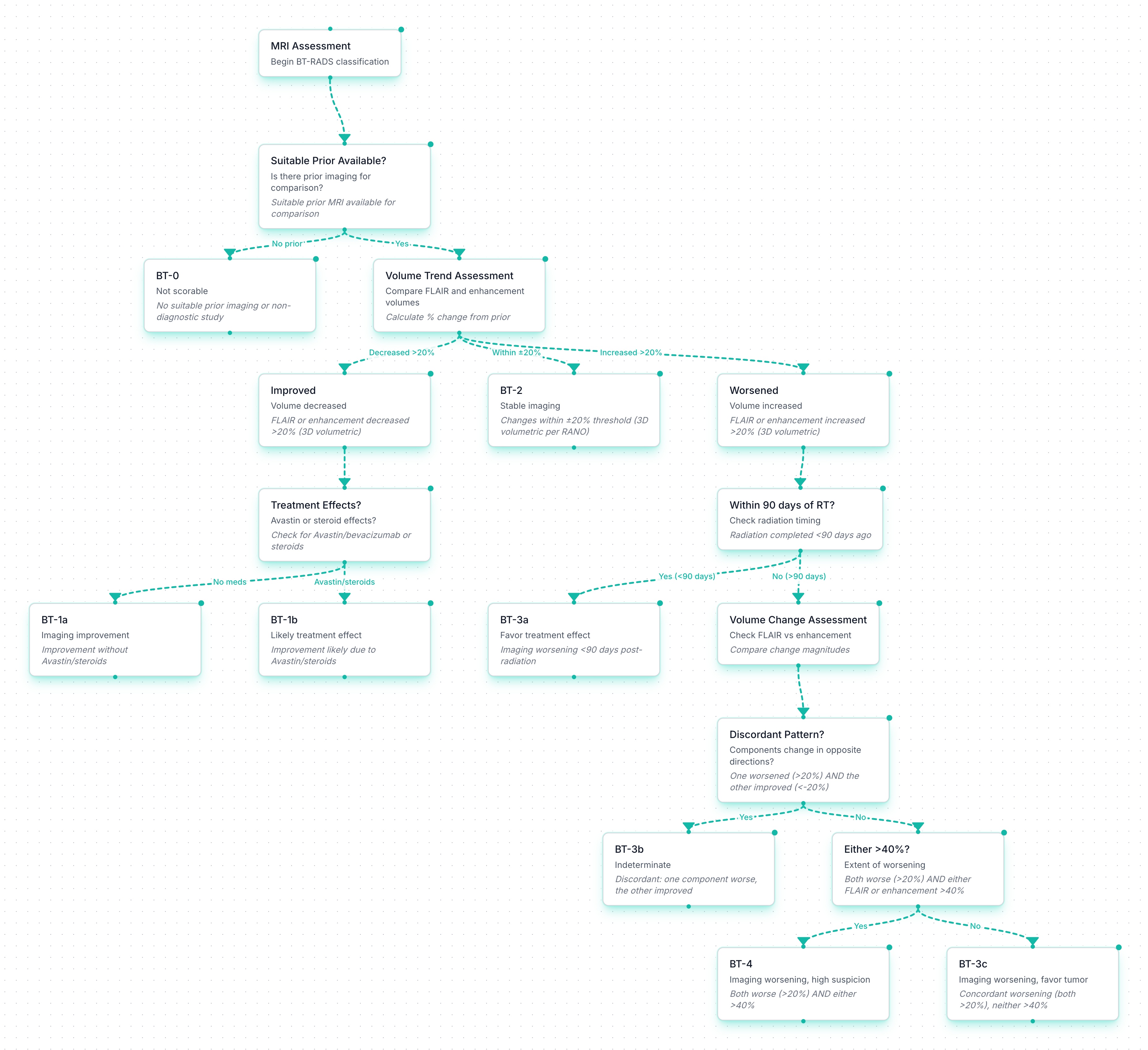}
\caption{BT-RADS decision flowchart implemented by the scoring agent. Sequential rules classify cases based on volumetric thresholds ($\pm$20\% stability, $>$40\% major change), medication effects (bevacizumab, steroids), and radiation timing ($<$90 days post-completion). Terminal nodes correspond to BT-RADS categories BT-0 through BT-4.}
\label{fig:flowchart}
\end{figure}

The multi-agent LLM system employs three specialized agents---an orchestrator agent, an extractor agent, and a scorer agent (Figure~\ref{fig:architecture}). This architecture separates extraction from classification into modular components. The extractor agent processes clinical notes to identify steroid status, bevacizumab status, and radiation therapy completion date. Schema-constrained generation with Pydantic validation ensures outputs conform to predefined categorical formats (e.g., steroid status must be ``active,'' ``recent,'' or ``none''). The scorer agent implements BT-RADS decision logic through sequential rule application (Figure~\ref{fig:flowchart}), integrating extracted clinical variables with volumetric measurements. Each extraction includes an evidence span---the specific text passage supporting the determination---enabling verification of the system's reasoning.

\subsection{Reference Standard and Validation}

Initial clinical BT-RADS classifications were obtained from radiology reports generated during routine clinical workflows by radiologists with varying degrees of training and expertise (including both fellowship trained neuroradiologists and non-neuroradiologists). Prior work demonstrated substantial discrepancies in clinical BT-RADS scores assigned during routine clinical workflow~\cite{zhang2025} with examples including comparison to an incorrect baseline study, non-standard scoring (e.g. BT-3 without a letter designation), and failure to identify relevant clinical variables such as bevacizumab therapy. To establish a rigorous reference standard, all 509 cases were independently re-scored by a single board-certified neuroradiologist with subspecialty expertise in neuro-oncology imaging, who reviewed the relevant clinical notes and MRI exams and applied the published BT-RADS flowchart and scoring criteria~\cite{btrads2026,gore2019}. Three cases were assigned non-standard scores (e.g. BT-1 or BT-3 without subcategory specification) as available clinical and imaging information presented conflicting or subthreshold findings that precluded definitive subcategory determination.

\subsection{Statistical Analysis}

Classification accuracy was calculated as correct classifications divided by total evaluable cases. Wilson score 95\% confidence intervals were computed. McNemar's test with continuity correction compared system versus initial assessment accuracy. Cohen's kappa (unweighted and quadratic weighted) quantified agreement with expert annotated reference standard. Per-category sensitivity and one-versus-all diagnostic metrics (sensitivity, specificity, positive predictive value, negative predictive value, likelihood ratios) were calculated. Root cause analysis categorized misclassifications by error source. Performance ceiling analysis calculated theoretical accuracy accounting reference standard annotation exclusions, extraction errors, or both.

%% ============================================================
%% RESULTS
%% ============================================================
\section{Results}

\subsection{Study Cohort}

Of 509 examinations, 492 were evaluable after excluding 17 cases: 9 lacked suitable baseline imaging for longitudinal comparison (first scan at institution or interval exceeding 6 months), and 8 had documented segmentation quality issues identified during quality assurance review. Among the 492 evaluable examinations, patient age mean and SD were 57.1 $\pm$ 13.1 years (median 59; IQR 49--67; range 19--88), with 280 (57.5\%) of the patients being male. Median baseline-to-follow-up scan interval was 63 days (IQR, 56--83; range, 9--1575). FLAIR volume changes ranged from $-$87\% to $+$854\% (median $+$8\%; IQR, $-$8\% to $+$41\%). Enhancement volume changes ranged from $-$100\% to $+$2882\% (median 0\%; IQR, $-$18\% to $+$46\%) (Supplementary Figure S1). Expert reference standard distribution (Table~\ref{tab:distribution}) was: BT-2 (stable) 156/492 (31.7\%), BT-3c 115/492 (23.4\%), BT-4 75/492 (15.2\%), BT-1a 55/492 (11.2\%), BT-1b 51/492 (10.4\%), BT-3b 21/492 (4.3\%), BT-3a 16/492 (3.3\%), and non-standard annotations 3/492 (0.6\%). The 3 non-standard cases were retained in the accuracy denominator; because the system predicts only standard BT-RADS categories, these cases were counted as misclassifications.

\begin{table}[H]
\centering
\caption{\textbf{Expert Reference Standard BT-RADS Distribution}}
\label{tab:distribution}
\begin{tabular}{llcc}
\toprule
\textbf{Category} & \textbf{Description} & \textbf{N} & \textbf{Percentage} \\
\midrule
BT-1a & Imaging improvement & 55 & 11.2\% \\
BT-1b & Likely treatment effect & 51 & 10.4\% \\
BT-2 & Stable imaging & 156 & 31.7\% \\
BT-3a & Favor treatment effect & 16 & 3.3\% \\
BT-3b & Indeterminate & 21 & 4.3\% \\
BT-3c & Favor tumor & 115 & 23.4\% \\
BT-4 & High suspicion for tumor & 75 & 15.2\% \\
Other & Non-standard annotation* & 3 & 0.6\% \\
\midrule
Total & & 492 & 100\% \\
\bottomrule
\end{tabular}

\vspace{0.5em}
{\footnotesize *Non-standard annotations include 3 cases (BT-1 or BT-3 without subcategory designation) where the expert neuroradiologist determined that available clinical and imaging information was insufficient to differentiate subcategories. $P$ value from McNemar test with continuity correction ($\chi^{2} = 28.62$). 95\% confidence intervals calculated using the Wilson score method. Classification accuracy calculated as agreement with expert reference standard.}
\end{table}

\subsection{Overall Classification Performance}

\begin{figure}[H]
\centering
\includegraphics[width=\textwidth]{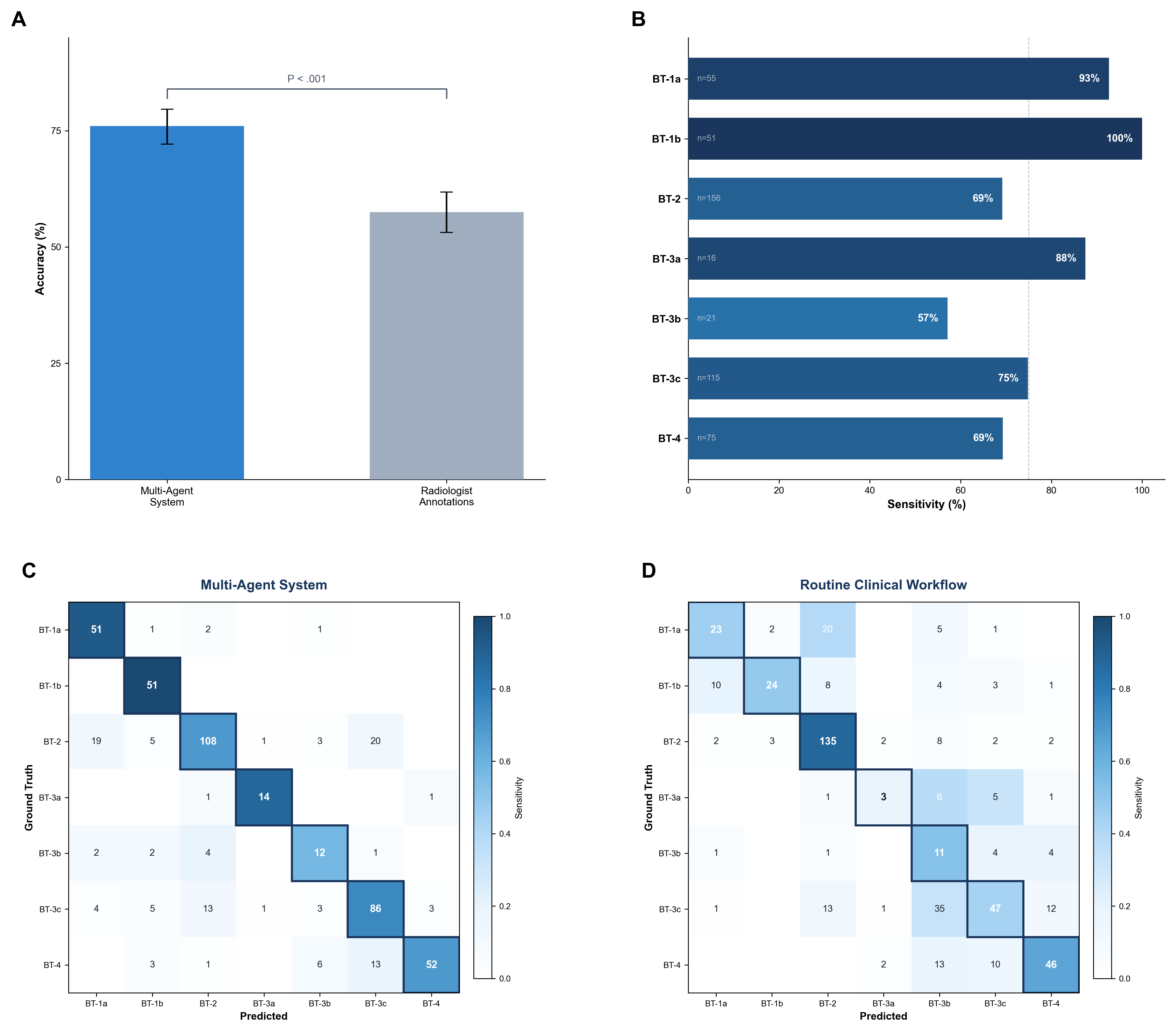}
\caption{Composite classification performance. (A) Overall accuracy: multi-agent system versus initial workflow (McNemar $P < .001$); error bars represent 95\% confidence intervals. (B) Per-category sensitivity of the agentic system in BT-RADS category order. (C) Multi-agent system confusion matrix. (D) Initial clinical assessment confusion matrix. In C and D, color intensity represents row-normalized sensitivity; cell values are raw counts; diagonal cells (correct classifications) are outlined.}
\label{fig:performance}
\end{figure}

The system achieved 374/492 (76.0\%; 95\% CI, 72.1\%--79.6\%) classification accuracy compared with 283/492 (57.5\%; 95\% CI, 53.1\%--61.8\%) for initial clinical interpretations (Figure~\ref{fig:performance}A). The absolute difference was 18.5 percentage points (McNemar $\chi^{2} = 28.62$, $P <.001$). Cohen's $\kappa$ was 0.708 (95\% CI, 0.667--0.750). Quadratic weighted $\kappa$ was 0.803 (95\% CI, 0.724--0.882).

\subsection{Per-Category Classification Performance}

Per-category sensitivity revealed a context-threshold pattern (Table~\ref{tab:sensitivity}, Figure~\ref{fig:performance}B). Categories requiring clinical context integration---BT-1a, BT-1b, and BT-3a---achieved high sensitivity (87.5\%--100\%), while categories determined primarily by volumetric thresholds---BT-2, BT-3c, and BT-4---showed moderate sensitivity (69.2\%--74.8\%). BT-3b had the lowest sensitivity at 12/21 (57.1\%), reflecting difficulty in cases where enhancement and FLAIR volume trends diverge. Subgroup analyses by temporal window, medication status, and enhancement magnitude are presented in Supplementary Table S1. The confusion matrix (Figure~\ref{fig:performance}C) showed misclassifications predominantly between adjacent categories: BT-2 misclassified as BT-3c in 20/156 (12.8\%) cases, BT-3c as BT-2 in 13/115 (11.3\%), and BT-4 as BT-3c in 13/75 (17.3\%).

\begin{table}[H]
\centering
\caption{\textbf{Per-Category Classification Sensitivity}}
\label{tab:sensitivity}
\begin{tabular}{llcccc}
\toprule
\textbf{Category} & \textbf{Type} & \textbf{N (RS)} & \textbf{Correct} & \textbf{Sensitivity} & \textbf{95\% CI} \\
\midrule
BT-1a & Context-dependent & 55 & 51 & 92.7\% & 82.7\%--97.1\% \\
BT-1b & Context-dependent & 51 & 51 & 100.0\% & 93.0\%--100.0\% \\
BT-2 & Threshold-dependent & 156 & 108 & 69.2\% & 61.6\%--75.9\% \\
BT-3a & Context-dependent & 16 & 14 & 87.5\% & 64.0\%--96.5\% \\
BT-3b & Threshold-dependent & 21 & 12 & 57.1\% & 36.5\%--75.5\% \\
BT-3c & Threshold-dependent & 115 & 86 & 74.8\% & 66.1\%--81.8\% \\
BT-4 & Threshold-dependent & 75 & 52 & 69.3\% & 58.2\%--78.6\% \\
\bottomrule
\end{tabular}

\vspace{0.5em}
{\footnotesize N (RS) = number of cases with expert reference standard annotation in each category. Context-dependent categories require extraction and integration of clinical variables (medication status, radiation timing). Threshold-dependent categories are determined primarily by volumetric threshold interpretation. 95\% confidence intervals calculated using the Wilson score method.}
\end{table}

\subsection{Diagnostic Performance}

One-versus-all diagnostic metrics for BT-4 (Table~\ref{tab:diagnostic}): sensitivity 52/75 (69.3\%), specificity 410/414 (99.0\%), positive predictive value 52/56 (92.9\%), negative predictive value 410/433 (94.7\%), positive likelihood ratio 69.3, negative likelihood ratio 0.31.

\begin{table}[H]
\centering
\caption{\textbf{Diagnostic Performance by BT-RADS Category}}
\label{tab:diagnostic}
\begin{tabular}{lcccccccc}
\toprule
\textbf{Category} & \textbf{N (RS)} & \textbf{Sensitivity} & \textbf{Specificity} & \textbf{PPV} & \textbf{NPV} & \textbf{LR+} & \textbf{LR$-$} \\
\midrule
BT-1a & 55 & 92.7\% & 94.2\% & 67.1\% & 99.0\% & 16.0 & 0.08 \\
BT-1b & 51 & 100.0\% & 96.3\% & 76.1\% & 100.0\% & 27.0 & 0.00 \\
BT-2 & 156 & 69.2\% & 93.7\% & 83.7\% & 86.7\% & 11.0 & 0.33 \\
BT-3a & 16 & 87.5\% & 99.6\% & 87.5\% & 99.6\% & 218.8 & 0.13 \\
BT-3b & 21 & 57.1\% & 97.2\% & 48.0\% & 98.1\% & 20.4 & 0.44 \\
BT-3c & 115 & 74.8\% & 90.9\% & 71.7\% & 92.1\% & 8.2 & 0.28 \\
BT-4 & 75 & 69.3\% & 99.0\% & 92.9\% & 94.7\% & 69.3 & 0.31 \\
\bottomrule
\end{tabular}

\vspace{0.5em}
{\footnotesize One-vs-all analysis treats each category as positive class versus all others combined. PPV = positive predictive value; NPV = negative predictive value; LR+ = positive likelihood ratio; LR$-$ = negative likelihood ratio. Extraction accuracy calculated as agreement between system extraction and expert annotation. 95\% confidence intervals calculated using the Wilson score method.}
\end{table}

\subsection{Clinical Variable Extraction}

The extractor agent achieved high accuracy across all three clinical variables, ranging from 432/492 (87.8\%) for steroid status to 478/492 (97.2\%) for bevacizumab status, with radiation date extraction at 448/492 (91.1\%).

\subsection{Misclassification Analysis}

Of 118 misclassifications, root cause analysis (Supplementary Table S2, Supplementary Figure S2) attributed 52/118 (44.1\%) to volumetric threshold errors, where volumetric changes fell near the $\pm$20\% stability or 40\% worsening cutoffs and small measurement differences alter categorical assignment independent of segmentation accuracy---distinct from the 8 excluded cases with large-scale segmentation quality issues; 34/118 (28.8\%) to clinical variable extraction errors (incorrect identification of medication status or radiation completion date from clinical notes); 18/118 (15.3\%) to algorithm design limitations (e.g., a new small enhancing lesion that did not significantly change total enhancement volume, incorrectly classified as stable); and 14/118 (11.9\%) to reference standard ambiguity (cases where even expert review could not definitively assign a subcategory). When stratified by extraction accuracy, 84/118 (71\%) of misclassifications occurred despite correct extraction of all three clinical variables, and 34/118 (29\%) involved at least one extraction error (Supplementary Table S3). Performance ceiling analysis (Table~\ref{tab:ceiling}): with perfect extraction, theoretical accuracy was 78.5\%; with perfect algorithm logic, 82.8\%; with both perfected, 88.1\%. The theoretical maximum of 99.4\% represents the scenario where only irreducible errors remain (threshold boundary ambiguity, inherent scoring variability, and 3 non-standard ground truth annotations).

\begin{table}[H]
\centering
\caption{\textbf{Theoretical Performance Ceiling}}
\label{tab:ceiling}
\begin{tabular}{lccp{7cm}}
\toprule
\textbf{Scenario} & \textbf{Accuracy} & \textbf{Improvement} & \textbf{Description} \\
\midrule
Current system & 76.0\% & Baseline & Observed performance with all error sources \\
Perfect extraction & 78.5\% & +2.5 pp & All 3 clinical variable extractions match expert \\
Perfect algorithm & 82.8\% & +6.8 pp & Optimal rule-based logic for edge cases \\
Perfect extraction + algorithm & 88.1\% & +12.1 pp & Correct extractions + optimal classification \\
Theoretical maximum & 99.4\% & +23.4 pp & Only inherent categorical ambiguity remains \\
\bottomrule
\end{tabular}

\vspace{0.5em}
{\footnotesize pp = percentage points. ``Perfect extraction'' models the scenario where all three clinical variable extractions (steroid status, bevacizumab status, radiation therapy completion date) match expert annotation; the modest gain (+2.5 pp) reflects that many extraction-attributed errors co-occur with threshold boundary ambiguity that persists regardless of extraction accuracy. ``Perfect algorithm'' models optimal rule-based classification logic for all edge cases, including discordant volume patterns and threshold boundary decisions. ``Perfect extraction + algorithm'' combines both improvements; the remaining 11.9\% gap to theoretical maximum reflects cases where categorical ambiguity is inherent to the BT-RADS framework. ``Theoretical maximum'' (99.4\%) assumes only the 3 cases with non-standard expert annotations remain unclassifiable.}
\end{table}

\begin{figure}[H]
\centering
\includegraphics[width=\textwidth]{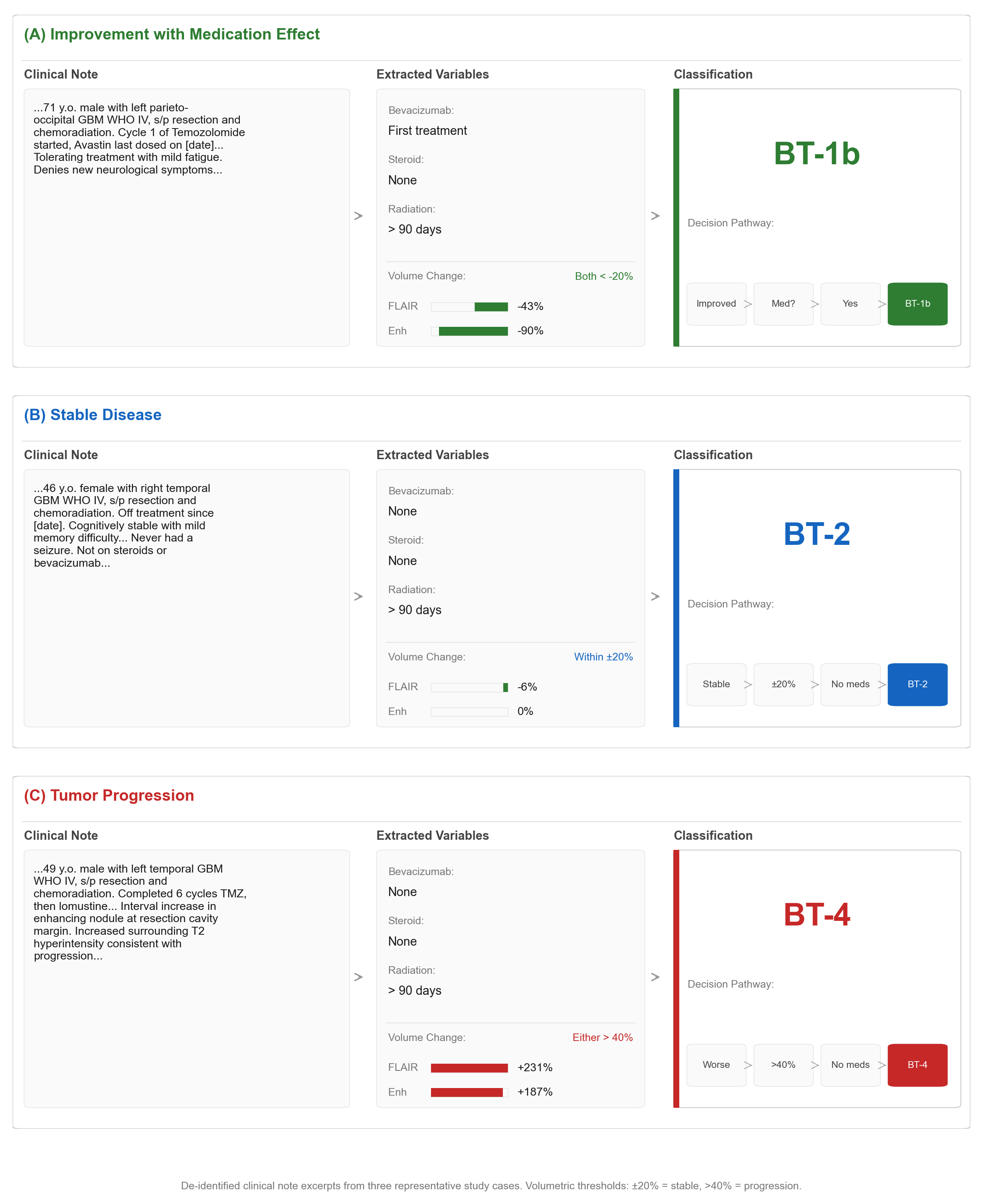}
\caption{Clinical information extraction and BT-RADS classification for three representative de-identified cases. (A) BT-1b: bevacizumab use identified from clinical notes; volumetric improvement routed through the medication effect pathway. (B) BT-2: both FLAIR and enhancement volumes stable within $\pm$20\%; no active medications. (C) BT-4: both components worsening (FLAIR $+$231\%, enhancement $+$187\%) with at least one exceeding 40\%; radiation completed more than 90 days prior. Each panel shows the clinical note excerpt with highlighted evidence spans, extracted variables, volumetric data, and the decision pathway leading to the final classification.}
\label{fig:cases}
\end{figure}

\subsection{Comparison with Initial Assessment}

\begin{figure}[H]
\centering
\includegraphics[width=\textwidth]{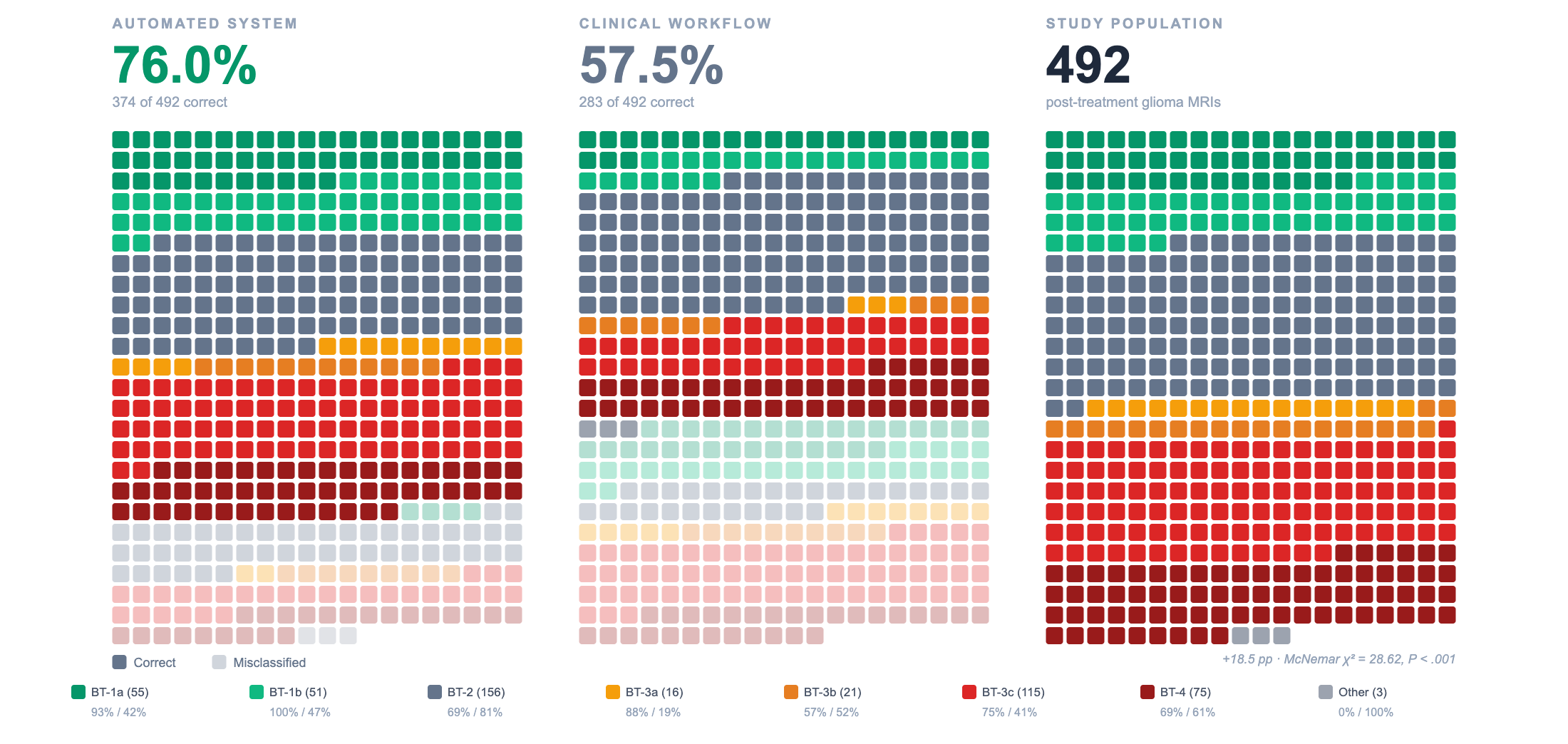}
\caption{Per-category classification concordance between the automated system (76.0\%; 374/492), initial clinical assessment (57.5\%; 283/492), and study population reference standard ($n = 492$). Each square represents one examination, colored by BT-RADS category. Solid squares indicate correct classification; faded squares indicate misclassification. Per-category accuracy is listed below (automated system / initial clinical assessment). McNemar $P < .001$.}
\label{fig:concordance}
\end{figure}

Classification concordance analysis (Figure~\ref{fig:concordance}): 187 cases (38.0\%) were correctly classified by both the automated system and original clinical assessment, 187 cases (38.0\%) were correctly classified only by the automated system, 96 cases (19.5\%) were correctly classified only by original clinical assessment, and 22 cases (4.5\%) were misclassified by both.

Analysis of initial BT-RADS classification discrepancies (Figure~\ref{fig:performance}D) revealed systematic error patterns. Of clinical annotations differing from expert reference standard: 27/492 (5.5\%) failed to identify medication effects (bevacizumab or steroids) affecting the BT-1a versus BT-1b distinction, and 13/492 (2.6\%) failed to correctly apply post-radiation window criteria for cases within 90 days of treatment completion. Additionally, 33/492 (6.7\%) of clinical annotations used non-standard BT-RADS nomenclature not defined in the official scoring system, including invalid subcategories (e.g., ``2b''; $n=13$), and categories without required subcategory specification (e.g., ``1'' or ``3'' without a/b/c designation; $n=20$). These findings highlight common pitfalls in manual BT-RADS application that the automated system is designed to avoid through systematic protocol adherence.

%% ============================================================
%% DISCUSSION
%% ============================================================
\section{Discussion}

This study describes the development and evaluation of an end-to-end automated system for BT-RADS classification of post-treatment brain gliomas, integrating deep-learning-based automated brain tumor volumetrics, LLM-based clinical variable extraction, and algorithmic decision logic. The system achieved overall accuracy comparable to previously reported human interrater reliability, with very high accuracy for several BT-RADS subcategories. Automated BT-RADS scores were also more consistent with the expert reference standard scoring compared to initial clinical scoring. In practice, such a system could pre-populate BT-RADS classifications by integrating imaging volumes with medication and radiation data extracted from clinical notes, allowing the interpreting radiologist to verify the extracted variables and proposed category rather than performing the complete multistep assessment de novo---reducing interpretive variability and the cognitive burden of synthesizing disparate data sources.

A context-threshold performance pattern was observed: the system achieved high sensitivity for categories requiring clinical context integration (BT-1a 92.7\%, BT-1b 100\%, BT-3a 87.5\%) and moderate sensitivity for threshold-dependent categories (BT-2 69.2\%, BT-3c 74.8\%, BT-4 69.3\%). Notably, BT-3a (87.5\%) exceeded BT-2 (69.2\%) sensitivity because BT-3a depends on radiation timing---a clinical context variable---whereas BT-2 depends on volumetric thresholds (Table~\ref{tab:sensitivity}). Root cause analysis attributed 52/118 (44.1\%) of errors to threshold boundary challenges, 34/118 (28.8\%) to extraction errors, 18/118 (15.3\%) to algorithm limitations, and 14/118 (11.9\%) to reference standard ambiguity. Threshold boundary errors reflect the inherent challenge of categorizing continuous volumetric data into discrete categories; additionally, volumetric measurement uncertainty due to potential over- or under-segmentation may contribute to misclassifications near these boundaries, as even validated segmentation algorithms exhibit variability that can shift measurements across categorical cutoffs. The remaining errors represent opportunities for improvement through enhanced prompts, alternative data sources, and refined decision logic.

The concentration of errors at volumetric threshold boundaries warrants consideration. Cases near the 20\% stability or 40\% worsening thresholds present challenges for both automated systems and human interpreters, as measurement variability and biological heterogeneity create inherent uncertainty. When the system classifies a case near a threshold differently from an expert, both interpretations may be clinically reasonable---consistent with prior interrater reliability studies demonstrating that BT-RADS achieves good but not perfect agreement (Gwet index 0.83) with disagreements concentrated at adjacent category boundaries~\cite{essien2024}.

The comparable performance of the automated system (76\%) and human interrater reliability ($\sim$80\%)~\cite{essien2024} suggests that automated BT-RADS classification is achievable with high accuracy, with remaining disagreements reflecting clinical complexity at threshold boundaries and outlier cases. Targeted improvements in extraction accuracy, particularly for steroid status (87.8\%), and refinement of algorithm logic may yield incremental gains~\cite{thirunavukarasu2023,singhal2023}.

The design supports clinical deployment through evidence-span linking for extraction verification, schema-constrained generation to prevent malformed responses~\cite{keshavarz2024,artsi2025}, and deterministic BT-RADS decision logic for consistent scoring. Cross-checks between extracted variables and volumetrics flag conflicts, and the system enforces valid BT-RADS nomenclature while systematically accounting for bevacizumab and steroid effects. Performance reflects a specific LLM configuration (GPT-oss; temperature 0.0; schema constraints) and may not fully generalize to other implementations.

Comparison with prior work provides context. Lee and colleagues achieved F1 scores of 0.68--0.72 for NLP-based BT-RADS inference from report text~\cite{lee2020}, but required an existing radiologist report. Zhang and colleagues showed that volumetric analysis alone achieved only moderate performance without subcategory differentiation, highlighting the necessity of clinical context integration beyond pure imaging quantification~\cite{zhang2025}. The present system extends these approaches by integrating volumetrics with explicit algorithmic logic and clinical variable extraction, accounting for medication effects and radiation timing. Observed extraction accuracies (87.8\%--97.2\%) align with prior open-weight LLM studies~\cite{leguellec2024,jabal2025} and larger-scale evaluations~\cite{li2025}.

These results support use as collaborative clinical decision support rather than fully autonomous scoring. Agreement with expert reference was substantial (Cohen's $\kappa$ = 0.708) and excellent when accounting for ordinality (quadratic weighted $\kappa$ = 0.803), with most errors between adjacent categories. High BT-4 specificity (99.0\%) and positive predictive value (92.9\%) indicate reliable progression detection. Local open-weight inference (GPT-oss, 20B) reduces privacy and cost barriers. The asymmetric discordance pattern---more cases correctly classified only by the system versus only by initial assessment---suggests complementary error modes amenable to combined approaches.

Limitations include single-institution retrospective design limiting generalizability, dependence on upstream segmentation quality (over- or under-segmentation contributes to threshold boundary misclassifications), and aggregate volumetric assessment that may mask clinically significant changes in individual lesions. The current approach computes total volumes without lesion-wise analysis; connected component analysis could enable detection of new discrete lesions and tracking of individual tumor foci---potentially improving accuracy for complex heterogeneous cases. The BT-3b category presented particular difficulty when volumetric components exhibit opposite trends, where the enhancement priority rule did not fully align with expert interpretation. Future directions include multi-institutional validation, prospective comparison with standard-of-care reporting, confidence metrics for uncertain classifications, and evaluation of whether differences between human and automated classifications correlate with patient survival outcomes.

In conclusion, a multi-agent LLM system achieved significantly higher BT-RADS classification accuracy than initial clinical assessments with high sensitivity for context-dependent categories (87.5\%--100\%) and high positive predictive value for BT-4 detection (92.9\%). The system performed comparably to reported human interrater reliability, with errors concentrated at volumetric threshold boundaries where classification uncertainty is anticipated. Evidence-span linking enables verification of extraction provenance, supporting deployment as clinical decision support. Future improvements in extraction accuracy and algorithm refinement may yield additional gains.

%% ============================================================
%% DATA AVAILABILITY
%% ============================================================
\section*{Data Availability}

The classification system source code is available upon request for academic research.

%% ============================================================
%% REFERENCES
%% ============================================================

%% ============================================================
%% SUPPLEMENTARY MATERIALS
%% ============================================================
\newpage
\appendix
\section*{Supplementary Materials}

\subsection*{Supplementary Tables}

\begin{table}[H]
\centering
\caption*{\textbf{Table S1. Characteristics of Excluded Cases Compared to Evaluable Cohort}}
\vspace{0.3em}
\begin{tabular}{lcc}
\toprule
\textbf{Characteristic} & \textbf{Excluded ($n=17$)} & \textbf{Evaluable ($n=492$)} \\
\midrule
\textit{Exclusion Reason} & & \\
\quad No suitable baseline & 9 (52.9\%) & --- \\
\quad Segmentation quality issue & 8 (47.1\%) & --- \\
\textit{Volumetric Changes} & & \\
\quad FLAIR \% (mean $\pm$ SD) & $+$24.1\% $\pm$ 72.3 & $+$29.5\% $\pm$ 95.0 \\
\quad Enhancement \% (mean $\pm$ SD) & $+$20.9\% $\pm$ 88.5 & $+$49.0\% $\pm$ 234.7 \\
\textit{Medication Status} & & \\
\quad Bevacizumab active & 7 (41.2\%) & 163 (33.1\%) \\
\quad Steroids active & 1 (5.9\%) & 18 (3.7\%) \\
\bottomrule
\end{tabular}
\vspace{0.3em}

{\footnotesize Per STROBE reporting guidelines, excluded cases were characterized to assess potential selection bias. All exclusions were determined prior to system evaluation based on predefined quality assurance criteria.}
\end{table}

\begin{table}[H]
\centering
\caption*{\textbf{Table S1. Comprehensive Subgroup Analysis}}
\vspace{0.3em}
\begin{tabular}{llccc}
\toprule
\textbf{Subgroup} & \textbf{Stratum} & \textbf{N} & \textbf{Accuracy} & \textbf{95\% CI} \\
\midrule
\textit{Temporal (Post-RT)} & $<$90 days & 40 & 72.5\% & 57.2\%--83.9\% \\
 & 90--180 days & 85 & 87.1\% & 78.3\%--92.6\% \\
 & $>$180 days & 280 & 75.7\% & 70.4\%--80.4\% \\
 & Unknown & 84 & 70.2\% & 59.8\%--79.0\% \\
\textit{Medication} & Bevacizumab active & 167 & 74.9\% & 67.8\%--80.8\% \\
 & Steroids only & 74 & 85.1\% & 75.3\%--91.5\% \\
 & No medication effect & 248 & 75.0\% & 69.3\%--80.0\% \\
\textit{Enhancement} & Improved ($<$$-$20\%) & 121 & 74.4\% & 65.9\%--81.3\% \\
 & Stable ($\pm$20\%) & 201 & 75.6\% & 69.2\%--81.0\% \\
 & Worse ($>$20\%) & 167 & 79.0\% & 72.3\%--84.5\% \\
\bottomrule
\end{tabular}
\vspace{0.3em}

{\footnotesize Analysis performed on 489 cases with standard BT-RADS category annotations (excludes 3 cases with non-standard annotations). Subgroups within each category are mutually exclusive. 95\% confidence intervals calculated using the Wilson score method.}
\end{table}

\begin{table}[H]
\centering
\caption*{\textbf{Table S2. Error Attribution Analysis ($n=118$ misclassifications)}}
\vspace{0.3em}
\begin{tabular}{lccl}
\toprule
\textbf{Error Category} & \textbf{N} & \textbf{\%} & \textbf{Remediability} \\
\midrule
Threshold boundary & 52 & 44.1\% & Irreducible \\
Extraction error propagation & 34 & 28.8\% & Remediable \\
Algorithm limitation & 18 & 15.3\% & Remediable \\
Ground truth ambiguity & 14 & 11.9\% & Irreducible \\
\midrule
Total & 118 & 100\% & \\
\bottomrule
\end{tabular}
\vspace{0.3em}

{\footnotesize Remediable errors ($n=52$, 44.1\%): extraction error propagation and algorithm limitations represent errors potentially addressable through system improvements. Irreducible errors ($n=66$, 55.9\%): threshold boundary challenges and ground truth ambiguity reflect inherent classification challenges.}
\end{table}

\begin{table}[H]
\centering
\caption*{\textbf{Table S3. Classification Accuracy Stratified by Extraction Success}}
\vspace{0.3em}
\begin{tabular}{lcccc}
\toprule
\textbf{Extraction Status} & \textbf{N} & \textbf{Correct} & \textbf{Accuracy} & \textbf{95\% CI} \\
\midrule
All 3 extractions correct & 380 & 296 & 77.9\% & 73.5\%--81.8\% \\
Any extraction incorrect & 109 & 78 & 71.6\% & 62.5\%--79.2\% \\
Radiation date incorrect & 43 & 28 & 65.1\% & 50.2\%--77.6\% \\
Steroid status incorrect & 59 & 44 & 74.6\% & 62.2\%--83.9\% \\
Bevacizumab incorrect & 14 & 11 & 78.6\% & 52.4\%--92.4\% \\
\bottomrule
\end{tabular}
\vspace{0.3em}

{\footnotesize Analysis performed on 489 cases with standard BT-RADS category annotations (excludes 3 cases with non-standard annotations). Rows for individual extraction types are not mutually exclusive.}
\end{table}

\subsection*{Supplementary Figures}

\begin{figure}[H]
\centering
\includegraphics[width=\textwidth]{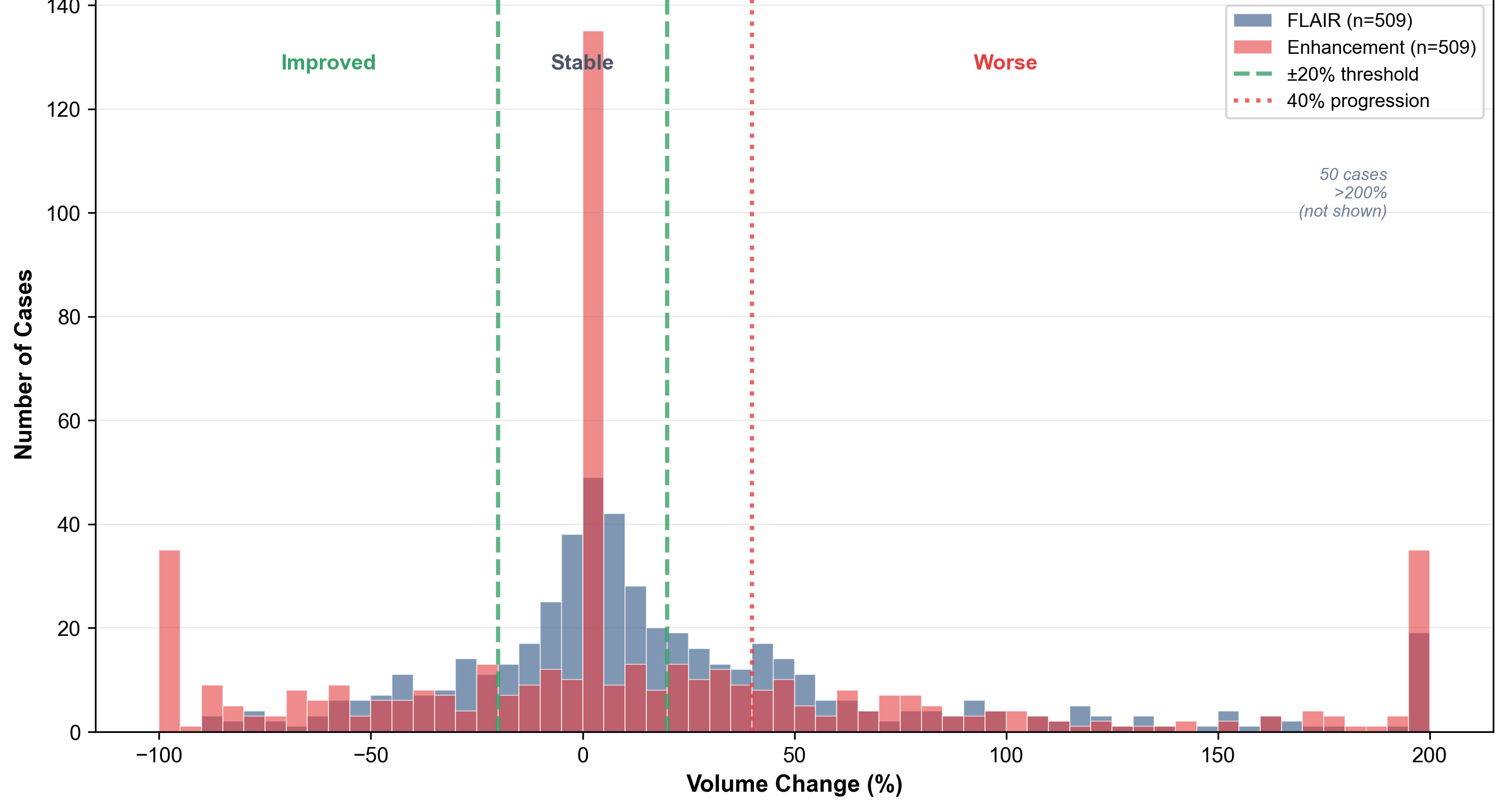}
\caption*{\textbf{Figure S1.} Distribution of Volumetric Changes: FLAIR and Enhancement ($n=509$). Overlaid histogram of FLAIR (blue) and enhancement (red) volumetric percentage changes across all 509 cases. Dashed lines indicate the $\pm$20\% stability and 40\% worsening thresholds used in BT-RADS classification.}
\end{figure}

\begin{figure}[H]
\centering
\includegraphics[width=\textwidth]{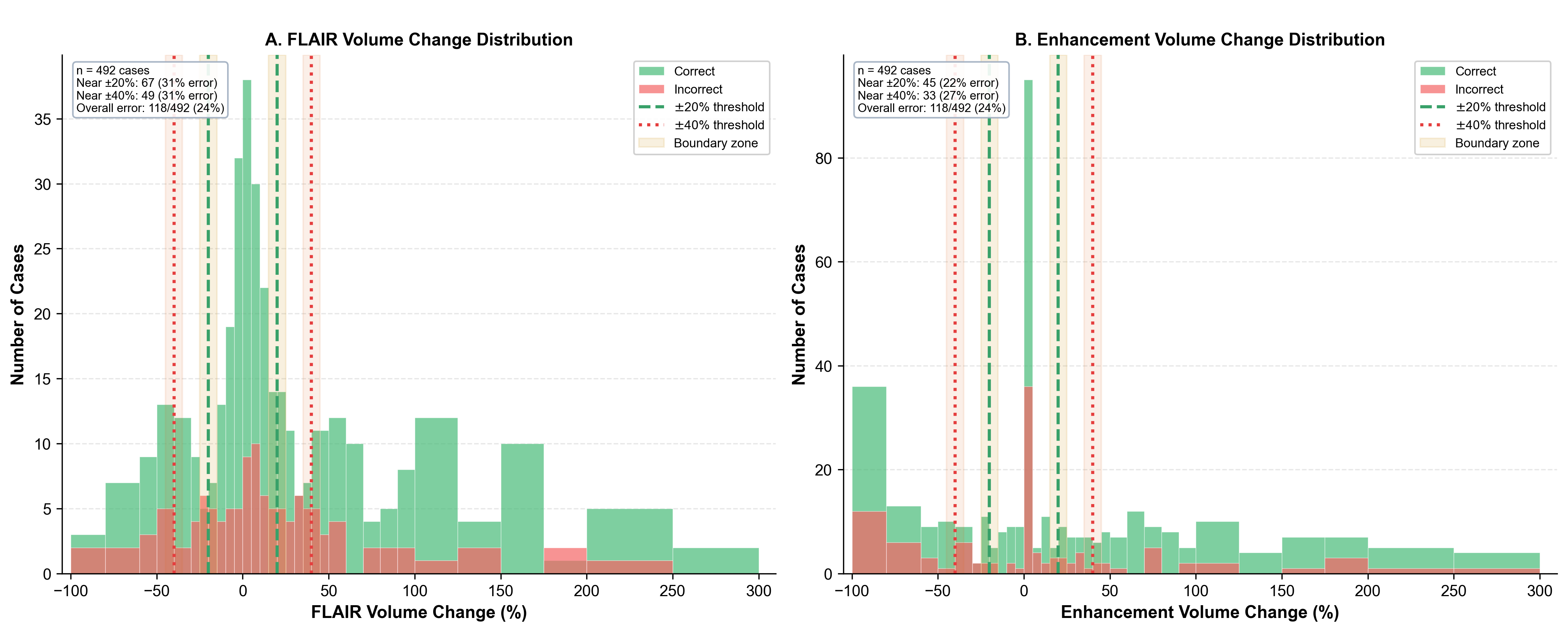}
\caption*{\textbf{Figure S2.} Threshold Boundary Analysis: Classification Accuracy by Volumetric Change Magnitude ($n=492$). Distribution of (A) FLAIR and (B) enhancement volumetric changes colored by classification outcome (green = correct, red = incorrect). Vertical dashed lines indicate BT-RADS decision thresholds.}
\end{figure}


\begin{thebibliography}{35}
\providecommand{\natexlab}[1]{#1}
\providecommand{\url}[1]{\texttt{#1}}
\expandafter\ifx\csname urlstyle\endcsname\relax
  \providecommand{\doi}[1]{doi: #1}\else
  \providecommand{\doi}{doi: \begingroup \urlstyle{rm}\Url}\fi

\bibitem[Ostrom et~al.(2021)Ostrom, Cioffi, Waite, Kruchko, and
  Barnholtz-Sloan]{ostrom2021}
Quinn~T. Ostrom, Gino Cioffi, Kristin Waite, Carol Kruchko, and Jill~S.
  Barnholtz-Sloan.
\newblock {CBTRUS} statistical report: primary brain and other central nervous
  system tumors diagnosed in the {United States} in 2014--2018.
\newblock \emph{Neuro-Oncology}, 23\penalty0 (12 Suppl 2):\penalty0
  iii1--iii105, 2021.

\bibitem[Tan et~al.(2020)Tan, Ashley, L{\'o}pez, Malinzak, Friedman, and
  Khasraw]{tan2020}
Aaron~C. Tan, David~M. Ashley, Giselle~Y. L{\'o}pez, Michael Malinzak, Henry~S.
  Friedman, and Mustafa Khasraw.
\newblock Management of glioblastoma: state of the art and future directions.
\newblock \emph{CA: A Cancer Journal for Clinicians}, 70\penalty0 (4):\penalty0
  299--312, 2020.

\bibitem[Brandsma et~al.(2008)Brandsma, Stalpers, Taal, Sminia, and van~den
  Bent]{brandsma2008}
Daphne Brandsma, Lukas Stalpers, Walter Taal, Peter Sminia, and Martin~J.
  van~den Bent.
\newblock Clinical features, mechanisms, and management of pseudoprogression in
  malignant gliomas.
\newblock \emph{The Lancet Oncology}, 9\penalty0 (5):\penalty0 453--461, 2008.

\bibitem[Abbasi et~al.(2018)Abbasi, Westerlaan, Holtman, Aden, Van~Laar, and
  Van Der~Hoorn]{abbasi2018}
Ali~Waez Abbasi, Henriette~E. Westerlaan, Gea~A. Holtman, Khadra~M. Aden,
  Peter~J. Van~Laar, and Anouk Van Der~Hoorn.
\newblock Incidence of tumour progression and pseudoprogression in high-grade
  gliomas: a systematic review and meta-analysis.
\newblock \emph{Clinical Neuroradiology}, 28\penalty0 (3):\penalty0 401--411,
  2018.

\bibitem[D'Orsi et~al.(2013)D'Orsi, Sickles, Mendelson, and Morris]{dorsi2013}
Carl~J. D'Orsi, Edward~A. Sickles, Ellen~B. Mendelson, and Elizabeth~A. Morris.
\newblock \emph{{ACR} {BI-RADS} Atlas: {Breast Imaging Reporting and Data
  System}}.
\newblock American College of Radiology, Reston, VA, 2013.

\bibitem[Weinberg et~al.(2018)Weinberg, Gore, Shu, et~al.]{weinberg2018}
Brent~D. Weinberg, Ashesh Gore, Hui-Kuo~G. Shu, et~al.
\newblock Management-based structured reporting of posttreatment glioma
  response with the brain tumor reporting and data system.
\newblock \emph{Journal of the American College of Radiology}, 15\penalty0
  (5):\penalty0 767--771, 2018.

\bibitem[{Brain Tumor Reporting and Data System (BT-RADS)}(2026)]{btrads2026}
{Brain Tumor Reporting and Data System (BT-RADS)}.
\newblock Scoring.
\newblock \url{https://btrads.com/scoring/}, 2026.
\newblock Accessed January 20, 2026.

\bibitem[Gore et~al.(2019)Gore, Hoch, Shu, Olson, Voloschin, and
  Weinberg]{gore2019}
Ashesh Gore, Michael~J. Hoch, Hui-Kuo~G. Shu, Jeffrey~J. Olson, Adam~D.
  Voloschin, and Brent~D. Weinberg.
\newblock Institutional implementation of a structured reporting system: our
  experience with the {Brain Tumor Reporting and Data System}.
\newblock \emph{Academic Radiology}, 26\penalty0 (6):\penalty0 974--980, 2019.

\bibitem[Essien et~al.(2024)Essien, Cooper, Gore, et~al.]{essien2024}
Mfoniso Essien, Michael~E. Cooper, Ashesh Gore, et~al.
\newblock Interrater agreement of {BT-RADS} for evaluation of follow-up {MRI}
  in patients with treated primary brain tumor.
\newblock \emph{AJNR American Journal of Neuroradiology}, 45\penalty0
  (8):\penalty0 1308--1315, 2024.

\bibitem[Zhu et~al.(2022)Zhu, Gilbert, Chetty, and Siddiqui]{zhu2022}
Sherry Zhu, Mark Gilbert, Indrin Chetty, and Farzan Siddiqui.
\newblock The 2021 landscape of {FDA}-approved artificial intelligence/machine
  learning-enabled medical devices: an analysis of the characteristics and
  intended use.
\newblock \emph{International Journal of Medical Informatics}, 165:\penalty0
  104828, 2022.

\bibitem[Jabal et~al.(2026)Jabal, Chisholm, Gupta, et~al.]{jabal2026}
Mohamed~Sobhi Jabal, Matthew Chisholm, Varun Gupta, et~al.
\newblock State and diffusion of {National Institutes of Health} funding of
  {AI} in radiology.
\newblock \emph{Journal of Imaging Informatics in Medicine}, 2026.
\newblock \doi{10.1007/s10278-026-01870-x}.
\newblock Published online February 11, 2026.

\bibitem[Prince et~al.(2024)Prince, Mirsky, Hankinson, and
  G{\"o}rg]{prince2024}
Eric~W. Prince, David~M. Mirsky, Todd~C. Hankinson, and Carsten G{\"o}rg.
\newblock Current state and promise of user-centered design to harness
  explainable {AI} in clinical decision-support systems for patients with {CNS}
  tumors.
\newblock \emph{Frontiers in Radiology}, 4:\penalty0 1433457, 2024.

\bibitem[Dietrich(2025)]{dietrich2025}
Nils Dietrich.
\newblock Agentic {AI} in radiology: emerging potential and unresolved
  challenges.
\newblock \emph{British Journal of Radiology}, 98\penalty0 (1174):\penalty0
  1582--1584, 2025.

\bibitem[Menze et~al.(2015)Menze, Jakab, Bauer, et~al.]{menze2015}
Bjoern~H. Menze, Andras Jakab, Stefan Bauer, et~al.
\newblock The multimodal brain tumor image segmentation benchmark ({BRATS}).
\newblock \emph{IEEE Transactions on Medical Imaging}, 34\penalty0
  (9):\penalty0 1993--2024, 2015.

\bibitem[Ghadimi et~al.(2025)Ghadimi, Vahdani, Karimi, et~al.]{ghadimi2025}
Delaram~J. Ghadimi, Amir~M. Vahdani, Hamed Karimi, et~al.
\newblock Deep learning-based techniques in glioma brain tumor segmentation
  using multi-parametric {MRI}: a review on clinical applications and future
  outlooks.
\newblock \emph{Journal of Magnetic Resonance Imaging}, 61\penalty0
  (3):\penalty0 1094--1109, 2025.

\bibitem[Diana-Albelda et~al.(2025)Diana-Albelda, Garcia-Martin, and
  Bescos]{diana2025}
Carlos Diana-Albelda, Alvaro Garcia-Martin, and Jes{\'u}s Bescos.
\newblock A review on deep learning methods for glioma segmentation,
  limitations, and future perspectives.
\newblock \emph{Journal of Imaging}, 11\penalty0 (7):\penalty0 269, 2025.

\bibitem[Gu et~al.(2025)Gu, Shao, Liao, et~al.]{gu2025}
Bijun Gu, Victor Shao, Zhiying Liao, et~al.
\newblock Scalable information extraction from free text electronic health
  records using large language models.
\newblock \emph{BMC Medical Research Methodology}, 25:\penalty0 23, 2025.

\bibitem[Reichenpfader et~al.(2024)Reichenpfader, Muller, and
  Denecke]{reichenpfader2024}
Daniel Reichenpfader, Henning Muller, and Kerstin Denecke.
\newblock A scoping review of large language model based approaches for
  information extraction from radiology reports.
\newblock \emph{npj Digital Medicine}, 7:\penalty0 222, 2024.

\bibitem[Bhayana(2024)]{bhayana2024}
Rajesh Bhayana.
\newblock Chatbots and large language models in radiology: a practical primer
  for clinical and research applications.
\newblock \emph{Radiology}, 310\penalty0 (1):\penalty0 e232756, 2024.

\bibitem[Li et~al.(2025)Li, Lacson, Guenette, et~al.]{li2025}
Kevin~W. Li, Ronilda Lacson, Jeffrey~P. Guenette, et~al.
\newblock Use of {ChatGPT} large language models to extract details of
  recommendations for additional imaging from free-text impressions of
  radiology reports.
\newblock \emph{AJR American Journal of Roentgenology}, 224\penalty0
  (4):\penalty0 e2432341, 2025.

\bibitem[Keshavarz et~al.(2024)Keshavarz, Bagherieh, Nabipoorashrafi,
  et~al.]{keshavarz2024}
Pedram Keshavarz, Sara Bagherieh, Seyed~Amirhossein Nabipoorashrafi, et~al.
\newblock {ChatGPT} in radiology: a systematic review of performance, pitfalls,
  and future perspectives.
\newblock \emph{Diagnostic and Interventional Imaging}, 105\penalty0
  (6--7):\penalty0 251--265, 2024.

\bibitem[Artsi et~al.(2025)Artsi, Sorin, Glicksberg, Korfiatis, Nadkarni, and
  Klang]{artsi2025}
Yaniv Artsi, Vera Sorin, Benjamin~S. Glicksberg, Panagiotis Korfiatis,
  Girish~N. Nadkarni, and Eyal Klang.
\newblock Large language models in real-world clinical workflows: a systematic
  review of applications and implementation.
\newblock \emph{Frontiers in Digital Health}, 7:\penalty0 1659134, 2025.

\bibitem[Lee et~al.(2025)Lee, Hadidchi, Coard, et~al.]{lee2025scoping}
Rachel~C. Lee, Ramin Hadidchi, Merissa~C. Coard, et~al.
\newblock Use of large language models on radiology reports: a scoping review.
\newblock \emph{Journal of the American College of Radiology}, 22\penalty0
  (10):\penalty0 101828, 2025.

\bibitem[Borkowski and Ben-Ari(2025)]{borkowski2025}
Andrew~A. Borkowski and Arik Ben-Ari.
\newblock Multiagent {AI} systems in health care: envisioning next-generation
  intelligence.
\newblock \emph{Federal Practitioner}, 42\penalty0 (5):\penalty0 188--194,
  2025.

\bibitem[Ke et~al.(2024)Ke, Yang, Lie, et~al.]{ke2024}
Yuhe Ke, Rui Yang, Sui~An Lie, et~al.
\newblock Mitigating cognitive biases in clinical decision-making through
  multi-agent conversations using large language models: simulation study.
\newblock \emph{Journal of Medical Internet Research}, 26:\penalty0 e59439,
  2024.

\bibitem[Liu et~al.(2025)Liu, Niu, Zhang, et~al.]{liu2025}
Fenglin Liu, Yongqi Niu, Qianqian Zhang, et~al.
\newblock A foundational architecture for {AI} agents in healthcare.
\newblock \emph{Cell Reports Medicine}, 6\penalty0 (9):\penalty0 102374, 2025.

\bibitem[Thirunavukarasu et~al.(2023)Thirunavukarasu, Ting, Elangovan,
  Gutierrez, Tan, and Ting]{thirunavukarasu2023}
Arun~James Thirunavukarasu, Daniel Shu~Wei Ting, Kabilan Elangovan, Laura
  Gutierrez, Ting~Fang Tan, and Daniel Shu~Wei Ting.
\newblock Large language models in medicine.
\newblock \emph{Nature Medicine}, 29\penalty0 (7):\penalty0 1930--1940, 2023.

\bibitem[Singhal et~al.(2023)Singhal, Azizi, Tu, et~al.]{singhal2023}
Karan Singhal, Shekoofeh Azizi, Tao Tu, et~al.
\newblock Large language models encode clinical knowledge.
\newblock \emph{Nature}, 620\penalty0 (7972):\penalty0 172--180, 2023.

\bibitem[Chen et~al.(2025)Chen, Yi, You, et~al.]{chen2025}
Xiangbin Chen, Haolei Yi, Mingwei You, et~al.
\newblock Enhancing diagnostic capability with multi-agents conversational
  large language models.
\newblock \emph{npj Digital Medicine}, 8\penalty0 (1):\penalty0 159, 2025.

\bibitem[Ferber et~al.(2025)Ferber, {El Nahhas}, W{\"o}lflein,
  et~al.]{ferber2025}
Daniel Ferber, Omar S.~M. {El Nahhas}, Georg W{\"o}lflein, et~al.
\newblock Development and validation of an autonomous artificial intelligence
  agent for clinical decision-making in oncology.
\newblock \emph{Nature Cancer}, 6\penalty0 (7):\penalty0 1337--1349, 2025.

\bibitem[Lee et~al.(2020)Lee, Weinberg, Gore, and Banerjee]{lee2020}
Soo~Jeon Lee, Brent~D. Weinberg, Ashesh Gore, and Imon Banerjee.
\newblock A scalable natural language processing for inferring {BT-RADS}
  categorization from unstructured brain magnetic resonance reports.
\newblock \emph{Journal of Digital Imaging}, 33:\penalty0 1393--1400, 2020.

\bibitem[{Le Guellec} et~al.(2024){Le Guellec}, Lefevre, Geay,
  et~al.]{leguellec2024}
Bastien {Le Guellec}, Antoine Lefevre, Charlotte Geay, et~al.
\newblock Performance of an open-source large language model in extracting
  information from free-text radiology reports.
\newblock \emph{Radiology: Artificial Intelligence}, 6\penalty0 (4):\penalty0
  e230364, 2024.

\bibitem[Jabal et~al.(2025)Jabal, Warman, Zhang, et~al.]{jabal2025}
Mohamed~Sobhi Jabal, Pranav Warman, Jikai Zhang, et~al.
\newblock Open-weight language models and retrieval-augmented generation for
  automated structured data extraction from diagnostic reports: assessment of
  approaches and parameters.
\newblock \emph{Radiology: Artificial Intelligence}, 7\penalty0 (3):\penalty0
  e240551, 2025.

\bibitem[Zhang et~al.(2025)Zhang, LaBella, Zhang, et~al.]{zhang2025}
Jikai Zhang, Dominic LaBella, Dylan Zhang, et~al.
\newblock Development and evaluation of automated artificial intelligence-based
  brain tumor response assessment in patients with glioblastoma.
\newblock \emph{AJNR American Journal of Neuroradiology}, 46\penalty0
  (5):\penalty0 990--998, 2025.

\bibitem[Ellingson et~al.(2015)Ellingson, Bendszus, Boxerman,
  et~al.]{ellingson2015}
Benjamin~M. Ellingson, Martin Bendszus, Jerrold Boxerman, et~al.
\newblock Consensus recommendations for a standardized brain tumor imaging
  protocol in clinical trials.
\newblock \emph{Neuro-Oncology}, 17\penalty0 (8):\penalty0 1188--1198, 2015.

\end{thebibliography}
\end{document}